\documentclass[conference]{IEEEtran}
\IEEEoverridecommandlockouts
\usepackage{cite}
\usepackage{amsmath,amssymb,amsfonts}
\usepackage{algorithmic}
\usepackage{graphicx}
\usepackage{textcomp}
\usepackage{xcolor}
\usepackage{booktabs}
\usepackage{times}
\usepackage{epsfig}
\usepackage{multirow}
\usepackage{stfloats}

\def\BibTeX{{\rm B\kern-.05em{\sc i\kern-.025em b}\kern-.08em
    T\kern-.1667em\lower.7ex\hbox{E}\kern-.125emX}}
\begin{document}

\title{Transferring Inter-Class Correlation}

\author{
Hui Wen \textsuperscript{\rm 1},
Yue Wu \textsuperscript{\rm 1},
Chenming Yang \textsuperscript{\rm 2},
Jingjing Li \textsuperscript{\rm 1},
Yue Zhu \textsuperscript{\rm 3},
Xu Jiang \textsuperscript{\rm 1},
Hancong Duan\textsuperscript{\rm 1} \\
\textsuperscript{\rm 1} School of Computer Science and Engineering, University of Electronic Science and Technology of China, China\\
\textsuperscript{\rm 2} National Key Lab on Communication, University of Electronic Science and Technology of China, China\\
\textsuperscript{\rm 3} NVIDIA Semiconductor Technology (Shanghai) Co., Ltd., Shanghai, China\\
Email:\{huiwen, chenmingyang\}@std.uestc.edu.cn, \{ywu,lijin117,jiangxu,duanhancong\}@uestc.edu.cn, \\tylerz@nvidia.com
}

\maketitle

\begin{abstract}
The Teacher-Student (T-S) framework is widely utilized in the classification tasks, through which the performance of one neural network (the student) can be improved by transferring knowledge from another trained neural network (the teacher).
Since the transferring knowledge is related to the network capacities and structures between the teacher and the student, how to define efficient knowledge remains an open question.
To address this issue, we design a novel transferring knowledge, the Self-Attention based Inter-Class Correlation (ICC) map in the output layer, and propose our T-S framework, Inter-Class Correlation Transfer (ICCT).
Notably, the analysis of our ICCT illustrates that the student combines with its own belief comprehensively adopting the teacher’s ICC map, rather than curtly mimic the teacher.
Since the ICC map can impose effective regularization, the knowledge from the teacher, which has a higher, equal, or lower capacity than the student can bring the benefit for the student training process.
Experimental results on the CIFAR-10, CIFAR-100, and ILSVRC2012 image classification datasets demonstrate that our ICCT can improve the student's performance and outperform other state-of-the-art T-S frameworks in T-S application scenarios with different network capacities and structures.
\end{abstract}


\section{Introduction}
\label{sec:introduction}

With the development of deep neural networks, considerable attractive approaches have been made to tackle challenging tasks.
Though numerous studies have been conducted to improve the model performance, the improvement which only relying on enlarging the number of parameters \cite{Large_Networks,krizhevsky2012imagenet} or searching for more complex network structures \cite{Inception-V4,Multigrid} becomes increasingly difficult.

\begin{figure*}[tbp]
    \centerline{\includegraphics[width=0.86\textwidth]{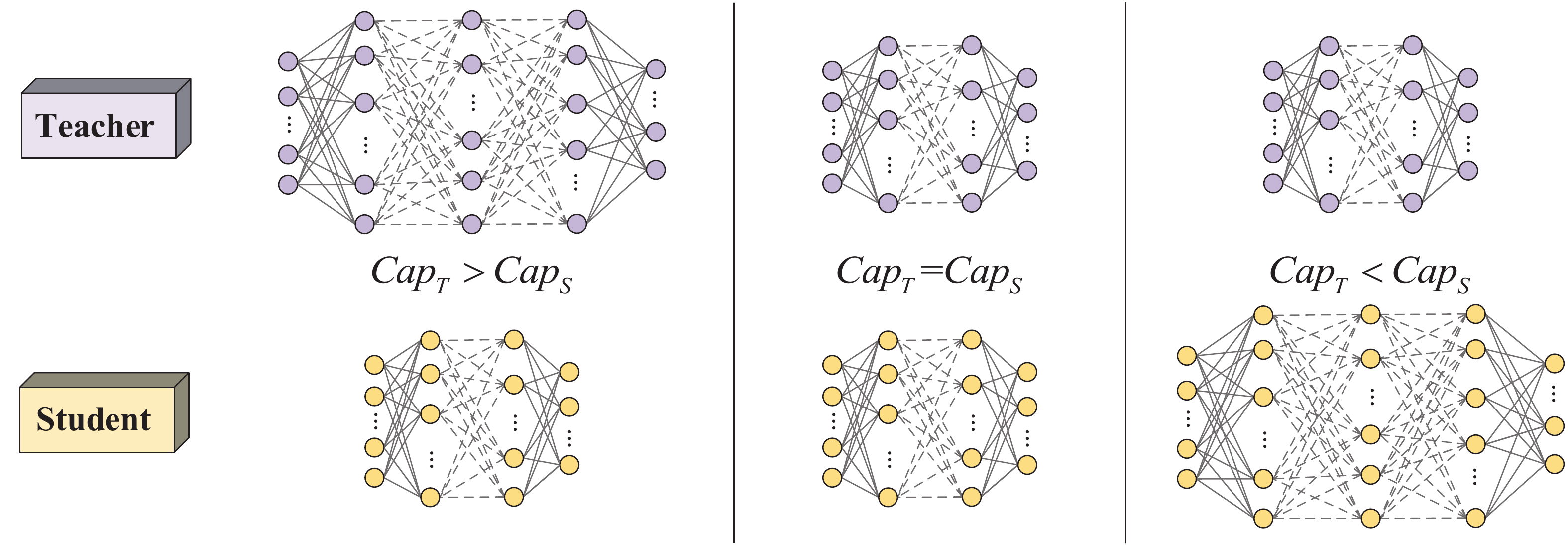}}
    \caption{
    Three capacity-based application scenarios ($Cap_{T}>Cap_{S}$, $Cap_{T}=Cap_{S}$, $Cap_{T}<Cap_{S}$).
    For clarity, we adopt the structure of ``MLP" as the demonstration.
    The purple is applied to mark the teacher, and the yellow is applied to mark the student.
    }
    \label{Application scenarios}
\end{figure*}

To overcome the difficulty, a supervised learning framework - Teacher-Student (T-S) framework - was proposed to promote the generalization accuracy of some neural network (the student) by bringing in transferring knowledge from another neural network (the teacher) \cite{bucilua2006model}.
Since the generalization accuracy can be regarded as the model capacity \cite{he2016deep}, we utilize the ``$Cap_{T}$" and ``$Cap_{S}$" to represent the capacity of the student and the teacher, respectively.
Furthermore, we employ the relationship between ``$Cap_{T}$" and ``$Cap_{S}$" to divide the T-S framework's application scenario.
In Figure \ref{Application scenarios}, we illustrate the three capacity-based T-S application scenarios.

The T-S framework was popular in the model compression area \cite{model_compression}, as for its capability of improving the
performance of the low-capacity student with few-parameters by a high-capacity teacher with large-parameters ($Cap_{T}>Cap_{S}$) \cite{Hinton2015DistillingTK, heo2019comprehensive, Similarity_Distillation}.
Recently, the T-S framework was proved to be also effective in improving the student performance in other capacity-based application scenarios, $Cap_{T}=Cap_{S}$ \cite{Born_Again,yuan2020revisiting} and $Cap_{T}<Cap_{S}$ \cite{Tolerant_Teacher,yuan2020revisiting}, in which the T-S framework excavates the potential generalization ability of the student.

Since T-S framework transfers knowledge from the teacher to the student, one of the critical issues is how to define the knowledge.
Hinton \emph{et al.} \cite{Hinton2015DistillingTK} first proposed the concept of Knowledge Distillation (KD) to guide the low-capacity student by a trained high-capacity teacher's softened output ($Cap_{T}>Cap_{S}$), controlled by a hyperparameter called ``temperature".
The transferring knowledge is only related to the categories of classification, and therefore does not constrain the network structure between the teacher and the student.
Furlanello \emph{et al.} \cite{Born_Again}, Yang \emph{et al.} \cite{Tolerant_Teacher}, and Yuan \emph{et al.} \cite{yuan2020revisiting} further exploited the intrinsic mechanism of the original KD and extended its capacity-based application scenarios where the student has equal ($Cap_{T}=Cap_{S}$) and higher capacity than the teacher ($Cap_{T}<Cap_{S}$).

Different from the knowledge definition in KD, another kind of transferring knowledge was defined in hidden layers.
For instance, Zagoruyko and Komodakis \cite{Paying_more_attention} defined the knowledge as the spatial attention map of the hidden-layers outputs.
Yim \emph{et al.} \cite{A_Gift} adopted the correlation matrix of two layers' feature maps as the knowledge.
Tung and Mori \cite{Similarity_Distillation} and Hou \emph{et al.} \cite{Lane_Detection} defined the knowledge by the similarity activation maps from hidden layers.
The advantage of these hidden-layers is that they can optimize students step by step.
However, since the feature maps of the student and the teacher are required to have the same shape \cite{Paying_more_attention, A_Gift, Lane_Detection}, these T-S frameworks with hidden-layers knowledge have to convert the hidden features.
Due to the missing of beneficial information during conversion processing, the performance of these frameworks is restricted by network structures between the teacher and the student.
Furthermore, the knowledge from hidden layers has only been explored in the $Cap_{T}>Cap_{S}$ scenario, and it is still unclear whether they are available in the $Cap_{T}=Cap_{S}$ or $Cap_{T}<Cap_{S}$ scenario in the image classification task.

\begin{figure}[tbp]
    \centerline{\includegraphics[width=0.48\textwidth]{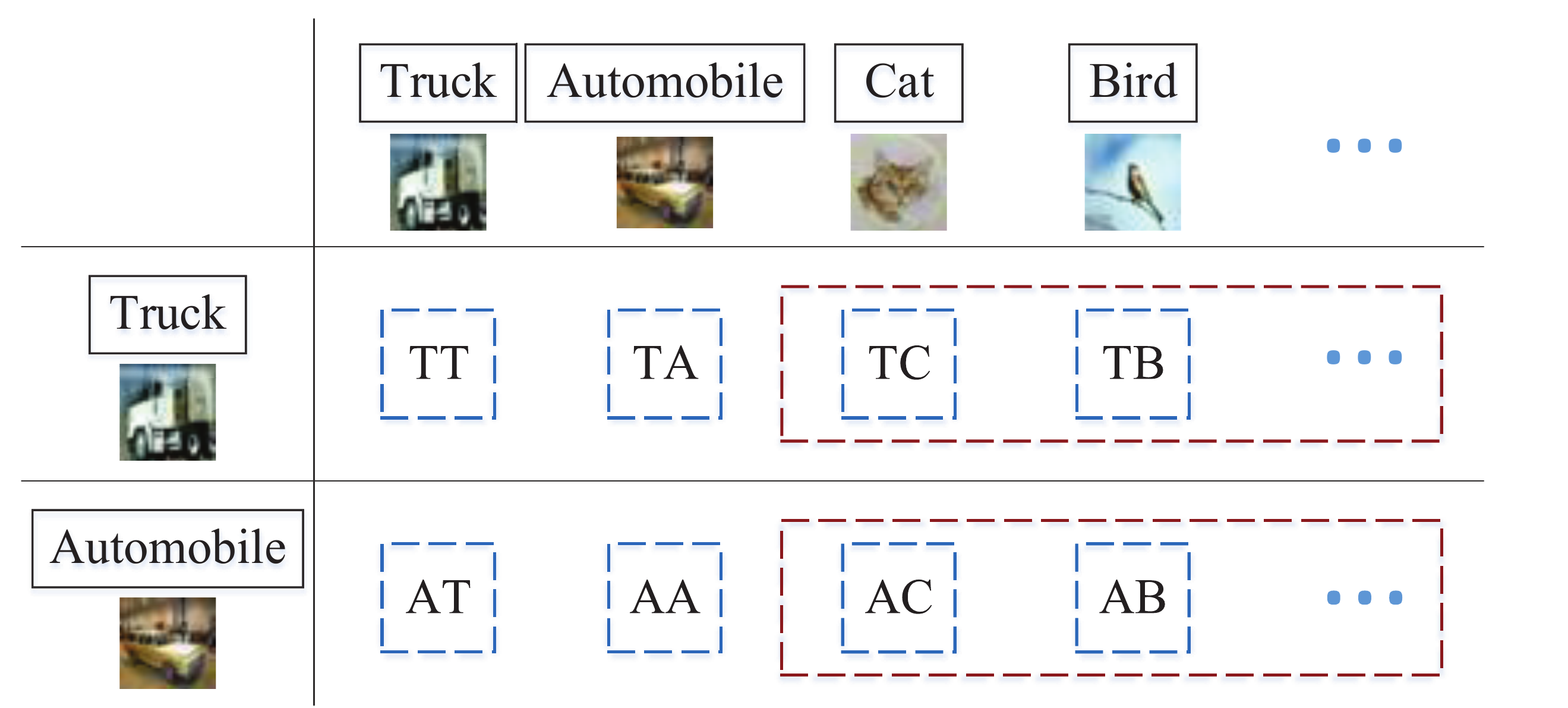}}
    \caption{
    Examples of pairwise Inter-Class Correlation in the CIFAR-10 dataset.
    ``TA" means the pairwise ICC between class ``Truck" and ``Automobile," and others are the same.
    Though the ``Truck" and the ``Automobile" are hard to distinguish, the ``TC" and ``TB" is different from ``AC" and ``AB."
    }
    \label{Classification of CIFAR-10}
\end{figure}

In this paper, we define knowledge based on a new perspective, named Inter-Class Correlation (ICC), for classification tasks.
Our intuitive motivation is that it is usually difficult to distinguish two similar classes if we can only obtain the information of these two classes.
However, if we can compare them with other dissimilar classes and form the ICC map, this task could be easier.
Figure \ref{Classification of CIFAR-10} shows an example of the ICC map in the CIFAR-10 dataset.
Though the ``Truck" and the ``Automobile" are similar, the similarity between the ``Truck" and other classes (like the ``Cat") can be different from the similarity between the ``Automobile" and the other classes (like the ``Cat").
Because the correlation of classes is only related to the images in the dataset, it is reasonable to require that the student and the teacher have similar ICC map corresponding with the same input image.

To form the ICC map of knowledge, we employ the popular Self-Attention mechanism \cite{vaswani2017attention} on every two ``logit" of the output layer.
We then use the Self-Attention based ICC map as the transferring knowledge and propose a novel T-S framework, named Inter-Class Correlation Transfer (ICCT).
Revealed by the analyzed gradients, the operating the student of our ICCT does not just mimic the teacher in that class.
Instead, the student gathers all the information related to this class in the ICC map and adds its own belief about this class.
We call this ``comprehensive" learning mode, which makes it reasonable to expect our ICCT to be applied in all three capacity-based application scenarios.

The ICC map is only related to classification outputs, like KD \cite{Hinton2015DistillingTK}, rather than the hidden features \cite{Paying_more_attention, A_Gift, Lane_Detection}.
Therefore, our ICCT eliminates the restrictions of network structure on the output shape of the hidden layer and avoids the loss of beneficial information from the teacher.
Notably, our ICC does not introduce hyperparameters, like ``temperature", which could change the original outputs.

To validate it, we design extensive experiments in three capacity-based T-S application scenarios $Cap_{T}>Cap_{S}$, $Cap_{T}=Cap_{S}$, and $Cap_{T}<Cap_{S}$, in which the structure of the teacher and student networks is unrestricted.
The experiments are conducted on three standard image classification datasets, CIFAR-10, CIFAR-100 \cite{krizhevsky2009learning}, and ILSVRC2012 \cite{russakovsky2015imagenet}, which have incremental complexity.
Experimental results demonstrate that our ICCT is not only able to improve the student's performance but also outperform KD and hidden-layers based T-S frameworks consistently, regardless of the application scenarios based on different network capacities and structures.

The main contributions of this paper are summarized as follows:
\begin{itemize}
  \item
  We design a novel T-S framework named Inter-Class Correlation Transfer (ICCT), which uses Self-Attention based Inter-Class Correlation as the transferring knowledge.
  Our ICCT eliminates the restrictions of network structure and avoids the loss of beneficial information from the teacher.
  \item
  We analyze the gradients of the ICC map and compare them with KD.
  The ICCT updates the student network in a ``comprehensive" mode, which means the student can comprehensively study teacher's knowledge based on its own belief, rather than completely mimicking the teacher.
  Benefited from the effective regularization of ICC, our ICCT can be flexibly implemented in different capacity-based T-S application scenarios.
  \item
  Experiments are conducted on three standard image classification datasets in three capacity-based application scenarios with different network structures to show the outstanding performance of our framework.
\end{itemize}



\section{preliminary}
\subsection{Inter-Class Correlation}
The Inter-Class Correlation has been well studied in clustering, measuring the relationship between categories among clusters \cite{wen2016discriminative}.
Benefited from the researches in clustering, studies have been conducted to improve the classification methods with the help of the Inter-Class Correlation.
In \cite{bengio2013using,rabinovich2007objects}, co-occurrence correlation was utilized to enforce object recognition in images.
Wen \emph{et al.} \cite{wen2016discriminative} proposed the ``center loss" as the loss function to get inter-class dispensation and intra-class compactness in face recognition.
Wu \emph{et al.} \cite{wu2014exploring} jointly learned feature relationships and exploited the inter-class relationships for improving video classification performance.

In recent years, another kind of correlation called Self-Attention has been attractive, since the success of ``Transformer" \cite{vaswani2017attention} in natural language processing.
The Self-Attention mechanism is a variant of the original attention mechanism, which is widely used in the Encoder-Decoder framework \cite{cho2014learning}.
Unlike the original attention mechanism extracting the correlations between the input and the output, the Self-Attention mechanism try to capture the correlations among different output neurons.
There are various operating modes to implement the Self-Attention mechanism, like dot product \cite{vaswani2017attention}, multiplicative \cite{paulus2017deep}, and additive \cite{bahdanau2014neural}.
To be noticed, the Self-Attention mechanism is served as a kind of intra-class correlations in natural language processing by treating the output as a whole.
However, in the classification task, the output neurons correspond with different classes, which means that the Self-Attention mechanism measures the Inter-Class Correlation.


\section{Inter-Class Correlation Transfer}
In this section, we present our Self-Attention based T-S framework, called Inter-Class Correlation Transfer (ICCT), which defines the ICC map as the transferring knowledge.
We first introduce the architecture of the T-S framework and then provide a Self-Attention based implementation of the ICC knowledge.
Based on the gradients, we finally explain how our ICCT works, which we called "comprehensive" mode.

\subsection{Teacher-Student Framework}
In the supervised image classification task, given a training set $\mathcal{D}= \{(\mathbf{x},\mathbf{y})|(\mathbf{x},\mathbf{y}) \in \mathcal{X} \times \mathcal{Y} \}$, where $\mathbf{x}$ and $\mathbf{y}$ are images and labels, the goal is to train a model $f(\mathbf{x}): \mapsto \mathcal{X} \times \mathcal{Y}$, which is able to generalize to unseen data.
In this paper, we qualify the model to be a neural network $f(\mathbf{x};\pmb{\theta})$ and learn the parameters $\pmb{\theta}$ via Empirical Risk Minimization (ERM).
The training process is to minimize the loss function:
\begin{equation}
  \pmb{\theta}^* = \arg \min_{{\pmb{\theta}}} \mathcal{L}_{label}(\mathbf{y},f(\mathbf{x};\pmb{\theta})),
\end{equation}
where $\mathcal{L}_{label}$ is some function to measure similarity.
Suppose that there are $N$ categories in total, and the output $\mathbf{y}'=f(\mathbf{x};\pmb{\theta}) = (y'_1,\ldots,y'_N)$ is generated by:
\begin{equation}
  y'_i = \frac{{e^{z_i}}}{{\sum\limits_{j = 1}^N {e^{z_j}} }},
\end{equation}
where $\mathbf{z}=(z_1,\ldots,z_N)$ is the input of the last softmax classifier and named by ``logits."
Due to the unbalance between model capacity and data space, this training process often limits the ability of the trained model generalizing to unseen data.

The T-S framework was proposed to tackle this problem.
In the T-S framework, the target model is treated as the student model $f(\mathbf{x};\pmb{\theta}_S)$.
Suppose we have got a trained teacher model $f(\mathbf{x};\pmb{\theta}_T)$.
We then need to define the knowledge transferred from $f(\mathbf{x};\pmb{\theta}_T)$ to $f(\mathbf{x};\pmb{\theta}_S)$.
One of the general definition of the knowledge is the one in Knowledge Distillation (KD) \cite{Hinton2015DistillingTK}, which gets a softened the softmax distribution $\mathbf{q} $ by distilling the logits $\mathbf{z}$:
\begin{equation}
  \label{KD}
   q_i = \frac{{e^{{{{z_i}} \mathord{\left/ {\vphantom {{{z_i}} M}} \right.
 \kern-\nulldelimiterspace} M}}}}{{\sum\limits_{j=1}^N {e^{{{{z_j}} \mathord{\left/
 {\vphantom {{{z_j}} M}} \right.
 \kern-\nulldelimiterspace} M}}} }},
\end{equation}
where $M$ is the soften hyperparameter, called the ``temperature."
Consequently, the loss function of the student model is:
\begin{equation}
  \begin{split}
  \pmb{\theta}_S^* &= \arg \min_{ {\pmb{\theta}_S}} (\mathcal{L}_{label}(\mathbf{y},f(\mathbf{x};\pmb{\theta}_S))+\lambda_{KD} \mathcal{L}_{KD}(\mathbf{q}_{S},\mathbf{q}_{T})),
\end{split}
\end{equation}
where $\mathcal{L}_{KD}$ is a similarity measure function, and $\lambda_{KD}$ controls the proportion of the two parts of loss.
Previous researches have demonstrated its ability to promote the generalization ability of the student by adding the teacher's knowledge in the student's loss function \cite{Hinton2015DistillingTK}.

Other hidden-layers based knowledge very likely restricts the output shape of the hidden layers, which is contrary to our demands.
So we omit the details of the hidden-layers based knowledge without losing integrity.

%

\begin{figure}[tbp]
    \centerline{\includegraphics[width=0.48\textwidth]{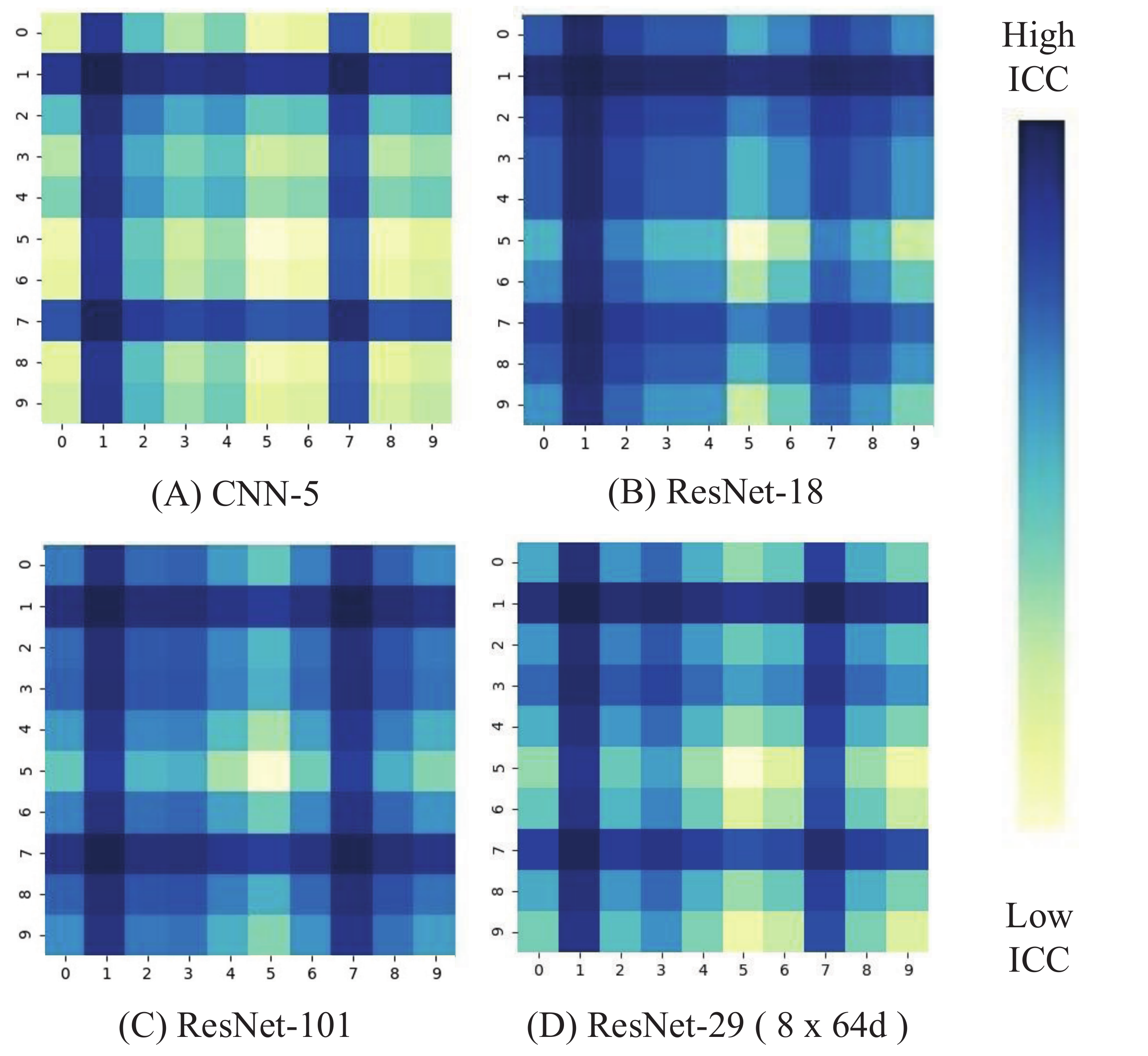}}
    \caption{
    The Self-Attention based ICC maps of the four networks with a same input image of the CIFAR-10 dataset.
    The structure of networks will explain in Section \ref{sec:E}.
    The row/column in the matrix contain all the pairwise ICC between a specific class and each class.
    Brighter colors indicate lower Inter-Class Correlation extents.
    }
    \label{Self-Attention matrix}
\end{figure}

\subsection{Self-Attention Based Inter-Class Correlation}
\label{sec:SA}
As illustrated in Section \ref{sec:introduction}, the ICC map can help the model to make better decisions in the image classification task.
To better utilize the information contained in the outputs, we propose a new definition of the transferring knowledge - Inter-Class Correlation (ICC) map, which contains the fine-grained relationship between every two classes in a mini-batch and represents a specific pattern of data.

%
%

The implementation of the ICC map is vital in our framework.
To simplify the computation and explain it clearly, we adopt the dot-product Self-Attention mechanism \cite{vaswani2017attention}.
The ICC map is calculated as follows.

Firstly, compute the unnormalized Self-Attention matrix by dot-product on ``logits" of the $s$-th sample in a mini-batch, $\mathbf{A}^s = [a^s_{ij}] = \mathbf{z}^s\mathbf{z}^{s\top} \in \mathbb{R}^{N \times N}$.
Then $\mathbf{A}^s$ should be normalized to make it comparable within different models and thus can be transferred.
As with the normalization in training single neural network, we choose the softmax function to generate the normalized form $\tilde{\mathbf{A}}^s=[\tilde{a}^s_{ij}] \in \mathbb{R}^{N \times N}$, where
\begin{equation}
  \label{aij}
  \begin{split}
    \tilde{a}^s_{ij} &= \frac{{e^{{a^s_{ij}}}}}{{\sum\limits_{u = 1}^N {\sum\limits_{v = 1}^N {e^{{a^s_{uv}}}} } }}
    =\frac{{e ^{ {{z^s_i}{z^s_j}} }}}{{\sum\limits_{u = 1}^N {\sum\limits_{v = 1}^N {e^ {{z^s_u}{z^s_v}}} } }}.
  \end{split}
\end{equation}
Finally, the matrices of all the samples should be averaged to measure the ICC in the mini-batch,
\begin{equation}
  \tilde{\mathbf{A}} = \frac{1}{b} \sum\limits_{s=1}^b \tilde{\mathbf{A}}^s,
\end{equation}
where $b$ is the batch size.
An example of the ICC maps of four neural networks with the same input is shown in Figure \ref{Self-Attention matrix}.
Apparently, the shades of the rows/columns in each matrix are very different, revealing the difference of the ICC between different classes.
The four matrices of four neural networks are similar, which is consistent with our illustration that the ICC is highly related to the input.

Noteworthy, the original Self-Attention mechanism \cite{vaswani2017attention} has components ``key," ``query," and ``value."
In the ICC map, the ``key" and ``query" are the ``logits" on the one hand, which are just like the original mechanism.
On the other hand, the ``value" can be seemed as ``1", which means that each class is equally treated.

Similar to other T-S frameworks, the hypothesis of our ICCT is that if an input image has a specific ICC map in the teacher network, it should have a similar ICC map in the student network.
Compared with KD, the ICC map $\tilde{\mathbf{A}}$ does not introduce additional hyperparameter ``temperature" as KD in Eq. (\ref{KD})
The reason is that we regard the ICC map as a representation of the average Inter-Class Correlation, which is determined by the samples in the mini-batch.
If we add the ``temperature" like KD, we change the correlations manually by reducing the higher correlations and amplifying the lower correlations.
The purpose of our ICCT is not to justify the ICC artificially, but to measure it as accurate as possible.



\begin{figure}[tbp]
    \centerline{\includegraphics[width=0.46\textwidth]{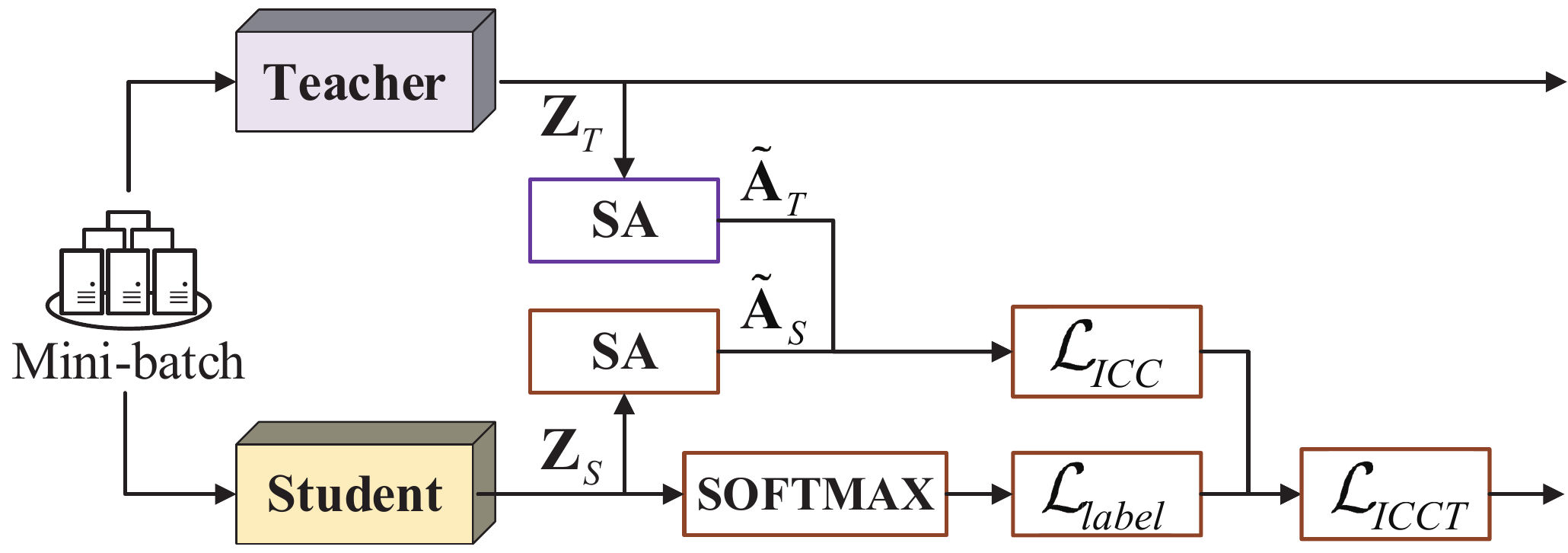}}
    \caption{
    ICCT Framework.
    This figure shows the training process of the student in our ICCT.
    The purple is applied to mark the teacher, and the yellow is applied to mark the student.
    }
    \label{ICCTFramework}
\end{figure}

After defining the ICC map as the transferring knowledge, we implement it in the T-S framework to form our ICCT.
Given an input mini-batch of $b$ images, we compute the ICC maps of the student $\tilde{\mathbf{A}}_S$ and the teacher $\tilde{\mathbf{A}}_T$ according to the corresponding ``logits."
The two matrices encode the average ICC of the input mini-batch.
The loss function of our ICCT is:
\begin{equation}
  \mathcal{L}_{ICCT} = \mathcal{L}_{label}(\mathbf{Y},f(\mathbf{X};\pmb{\theta}_S)) + \lambda_{ICC} \mathcal{L}_{ICC}(\tilde{\mathbf{A}}_T,\tilde{\mathbf{A}}_S),
\end{equation}
where $\mathbf{Y} = (\mathbf{y}^1,\ldots,\mathbf{y}^b)$ and $\mathbf{X} = (\mathbf{x}^1,\ldots,\mathbf{x}^b)$ are the labels and inputs in the mini-batch; $\lambda_{ICC}$ is a hyperparameter to balance the influence of labels and ICC maps.
In our experiments, we set $\mathcal{L}_{label}$ to be the cross-entropy function and $\mathcal{L}_{ICC}$ to be the KL-divergence between the teacher and The student.
The framework diagram is shown in Figure \ref{ICCTFramework}.

\subsection{Gradients of the ICCT}
In this section, we analyze the gradients with respect to the ``logits" to interpret how our ICCT works.
Suppose that there are $N$ classes in the dataset and $b$ samples in each mini-batch.

In our ICCT, the gradient of $\mathcal{L}_{ICC}$ with respect to the ``logits" are shown in Eq. (\ref{gradient of the batch-sample}):
\begin{equation}
  \begin{split}
    \label{gradient of the batch-sample}
    \frac{\partial \mathcal{L}_{ICC}}{\partial z_{k,S}^{s}}
    &= \frac{2}{b}\sum_{s=1}^{b}\sum\limits_{i=1}^N z_{i,S}^{s}(\tilde a_{ik,S}^{s} - \tilde a_{ik,T}^{s}),
  \end{split}
\end{equation}
where $\tilde{a}_{ik,S}^{s}$ and $\tilde{a}_{ik,T}^{s}$ are the corresponding element of the student's and the teacher's ICC maps
in the $s$-th sample.
The details of calculating the gradients are presented in the supplementary material.

To illustrate the different operating mode of ICCT, we compare its gradient with that of KD:
\begin{equation}
  \label{KDgradient}
  \begin{split}
  \frac{\partial \mathcal {L}_{KD}}{\partial {z}_{k,S}^{s}}&=\frac{1}{bM}\sum\limits_{s=1}^b(q_{k,S}^s - q_{k,T}^s)\\
  & =   \frac{1}{bM}\sum\limits_{s=1}^b(\frac{e^{(z_{k,S}^{s}/M)}}{\sum\limits_{j=1}^{N}e^{(z_ {j,S}^{s}/M)}}-\frac{e^{(z_{k,T}^{s}/M)}}{\sum\limits_{j=1}^{N}e^{(z_{j,T}^{s}/M)}}).
\end{split}
\end{equation}
Eq. (\ref{KDgradient}) reveals that the student's each ``logit" ${z}_{k,S}^{s}$ is updated independently by just mimicking the teacher's ``logit" ${z}_{k,T}^{s}$.
Each class is treated equally when minimizing the loss function.

In our ICCT, the student's each ``logit" ${z}_{k,S}^{s}$ is updated by gathering all the correlations between class $k$ and all the classes.
This could bring about more information to the student.
Moreover, due to the existence of ${z}_{i,S}^{s}$ in Eq. (\ref{gradient of the batch-sample}), the student does not equally treat all the classes.
Instead, the student can decide which class should be learned more, according to its own outputs.
${z}_{i,S}^{s}$ acts as an belief weight for the $i$-th class.
When ${z}_{i,S}^{s}$ is large, the student believes that class $i$ is possible to be the ground truth and tends to learn more about the gap in this class between the teacher and the student.
Otherwise, the student learns less about the gaps in the relatively less critical classes.
In such a way, the student can comprehensively study teacher’s knowledge based on its own belief.
We called this operating mode as ``comprehensive" mode, though which the student can learn smarter and may obtain better generalization capability.

\subsection{Differences With Previous Approaches}
Our ICCT defines the transferring knowledge at the output layer, similar to the traditional KD \ref{KD}.
Instead of using soften outputs as KD, we employ the Self-Attention mechanism to extract Inter-Class Correlation to form the knowledge.
Other hidden-layers based T-S frameworks \cite{Paying_more_attention,Similarity_Distillation,Lane_Detection} define the transferring knowledge at the hidden layers or blocks.

To be noticed, \cite{Similarity_Distillation,Lane_Detection} recently proposed activation-based T-S frameworks, which also utilize the concept of correlations.
However, there is a key difference between our ICCT and these two works.
In \cite{Similarity_Distillation,Lane_Detection}, the knowledge is defined by the similarity between two inputs' activation maps.
They hypothesize that if two inputs produce highly similar activations in the teacher network, it is beneficial to guide the student network towards a configuration that also results in the two inputs producing highly similar activations in the student.
However, in our ICCT, knowledge is the averaged Inter-Class Correlation in a mini-batch.
Our hypothesis is that the correlations between classes should be determined by the data.
The ICC maps of the student and the teacher are just two estimations about the correlations.
Our ICCT is to guide the student's estimation close to the teacher's estimation.

Moreover, \cite{Similarity_Distillation} mainly concentrated in the scenario $Cap_T>Cap_S$ for image classification, and \cite{Lane_Detection} considers the scenario $Cap_T=Cap_S$ for lane detection.
Our ICCT is also designed for the image classification task, but we consider more general scenarios  $Cap_T>Cap_S$, $Cap_T=Cap_S$, and $Cap_T<Cap_S$.
The $Cap_T=Cap_S$, and $Cap_T<Cap_S$ can further validate our hypothesis about the Inter-Class Correlation and benefit for promoting the performance of the student when we do not have trained high-capacity teachers.

\section{Experimental Evaluation}
\label{sec:E}

\subsection{Experimental Settings}

In this section, experiments are conducted to validate the efficiency of our ICCT in classification tasks, in which the median error of test over the ten standard splits is reported.

The experiments implement on three common datasets, CIFAR-10, CIFAR-100 \cite{krizhevsky2009learning}, and large-scale dataset ILSVRC2012 \cite{russakovsky2015imagenet}, and the complexity of these data increases successively.

To demonstrate the superiority of our ICCT, we compare the performance of commonly used frameworks Knowledge Distillation (KD) \cite{Hinton2015DistillingTK}, logit-transfer (LT) \cite{ba2014deep}, attention transfer (AT) \cite{Paying_more_attention}, and Similarity-Preserving (SP) \cite{Similarity_Distillation}, Tf-KD \cite{yuan2020revisiting} with our framework.

\begin{table*}[tb]
  \centering
  \renewcommand\tabcolsep{6.0pt}
  \fontsize{8}{12}\selectfont
  \caption{Parameters of networks.}
    \begin{tabular}{|c|c|c|c|c|c|c|c|c|c|}
    \hline
      Networks
      & CNN-5   & WRN-16-2   & WRN-28-2   & MobileNetV2
       & ResNet-18   & WRN-16-10   & ResNeXt-29  & WRN-28-10
       & ResNet-101 \cr\hline
      Parameters
      & 0.36M   & 0.70M             & 1.48M             & 3.42M
       & 11.17M       & 17.12M              & 34.43M     & 36.50M
       & 42.51M \cr\hline
    \end{tabular}
    \label{network parameters}
\end{table*}

\begin{table*}[t]
  \centering
  \renewcommand\tabcolsep{5.3pt}
  \fontsize{8}{12}\selectfont
  \caption{Test error (\%) in capacity-based application scenario $Cap_T > Cap_S$ on the CIFAR-100.
  ``S(B)" and ``T(B)" are short for ``the error rate of the baseline of the student" and ``the error rate of the baseline of the teacher," respectively.
  ``NA" means ``Not Available."}
  \begin{tabular}{|c|cc|c|cc|cccc|ccc|}
    \hline
      Heterogeneity   & S   & T   & PR   & S(B) & T(B)     & KD    & LT    & AT   & SP & \bf ICCT     & \bf ICCT+AT & \bf ICCT+SP    \cr\hline
      \multirow{3}{*}{Architecture}
        & CNN-5       & ResNeXt-29   & 1:95.64 & 39.66  & 17.70
          & 37.93  & 38.16  & NA     & 37.71   & 37.35  & 37.02  & 36.87 \cr
        & MobileNetV2 & ResNeXt-29   & 1:10.07 & 34.82  & 17.70
          & 33.78  & 33.95  & 33.26  & 32.81   & 31.84  & 31.43  & 31.12 \cr
        & ResNet-18   & ResNeXt-29   & 1:3.08  & 24.34  & 17.70
          & 23.35  & 23.32  & 23.17  & 22.98   & 22.32  & 21.94  & 21.45 \cr\hline
      \multirow{2}{*}{Depth}
        & ResNet-18   & ResNet-101   & 1:3.81  & 24.34  & 22.19
          & 23.52  & 23.59  & 23.43  & 23.19   & 22.41  & 22.15  & 21.76 \cr
        & WRN-16-10 & WRN-28-10      & 1:2.13  & 21.56  & 20.48
          & 21.10  & 21.22  & 20.97  & 20.84   & 20.53  & 20.36 & 20.05  \cr\hline
      Width
        & WRN-28-2 & WRN-28-10       & 1:24.66 & 25.28  & 20.48
          & 24.73  & 24.82  & 24.56  & 24.27   & 23.61  & 23.36  & 23.14 \cr\hline
      Depth \& Width
        & WRN-16-2 & WRN-28-10       & 1:52.14 & 27.29  & 20.48
          & 26.84  & 26.92  & 26.58  & 26.27   & 25.66  & 25.42  & 25.23 \cr\hline
          \bottomrule[0.5pt]
      Architecture
        & ResNet-101 & ResNeXt-29    & 1:0.81  & 22.19  & 17.70
          & 21.66  & 21.72  & 21.48  & 21.30   & 20.78  & 20.61  & 20.54  \cr\hline
      \end{tabular}
    \label{tab:CIFAR-100_T>S}
\end{table*}

\begin{table}[tb]
  \centering
  \renewcommand\tabcolsep{6.7pt}
  \fontsize{8}{12}\selectfont
  \caption{Test error (\%) in capacity-based application scenario ${Cap_T = Cap_S}$ on the CIFAR-100.}
    \begin{tabular}{|c|c|cccc|}
    \hline
      Model   & Generations  & KD      & LT     & Tf-KD  & \bf ICCT \cr\hline
      \multirow{5}{*}{ResNet-18}
              & Baseline     & 24.34   & 24.34  & 24.34  & 24.34   \\ \cline{2-6}
              & Gen \#1      & 23.97   & 23.83  & 23.15  & 22.90   \\ 
              & Gen \#2      & 23.62   & 23.96  & 23.32  & 22.97   \\ 
              & Gen \#3      & 23.59   & 23.68  & 23.49  & 22.74   \\ 
              & Gen \#4      & 23.84   & 23.63  & 23.22  & 23.08   \cr\hline
     \multirow{5}{*}{ResNet-101}
              & Baseline     & 22.19 	 & 22.19  & 22.19  & 22.19  \\ \cline{2-6}
              & Gen \#1      & 21.98   & 21.83  & 21.06  & 20.86  \\ 
              & Gen \#2      & 21.80   & 21.86  & 20.73  & 20.95  \\ 
              & Gen \#3      & 21.76   & 21.81  & 20.87  & 20.84  \\ 
              & Gen \#4      & 21.91   & 22.04  & 21.45  & 20.87  \cr\hline
     \multirow{5}{*}{MobileNetV2}
              & Baseline     & 34.82   & 34.82  & 34.82  & 34.82  \\ \cline{2-6}
              & Gen \#1      & 34.65   & 34.04  & 33.74  & 33.25 	\\ 
              & Gen \#2      & 34.12   & 34.15  & 33.68  & 32.07  \\ 
              & Gen \#3      & 33.90   & 34.28  & 32.87  & 32.64  \\ 
              & Gen \#4      & 34.18   & 34.47  & 33.02  & 32.92  \cr\hline
    \end{tabular}
    \label{tab:CIFAR-100_T=S}
\end{table}

\subsubsection{Experimental Networks}
To further illustrate that our ICCT is not constrained by the network structure, we utilize five kinds of networks with heterogeneous structures.

We start with a tiny toy model, a convolutional neural network with 5 layers (CNN-5).
We perform three 3 $\times$ 3 convolutional layers, each followed by batch normalization, max pooling, and ReLU activation.
The first convolutional layer has 32 output channels, while the following two convolutional layers have 64 and 128 output channels, respectively.
After that, two consecutive fully-connected layers are added.
We then choose four mainstream classification models, ResNet \cite{he2016deep}, ResNeXt \cite{xie2017aggregated}, WideResNet \cite{WideResNet}, and MobileNetV2 \cite{mobilenetv2}.
For the sake of brevity, we abbreviate ``WideResNet" as ``WRN."

Distillation between different types of the model structure has also been considered.
Based on the above five network structures, we design four kinds of pairs between the student and the teacher: heterogeneity of architecture, heterogeneity of depth, heterogeneity of width, and heterogeneity of depth\&width.


\subsubsection{Experimental Scenarios}
To show our ICCT can flexibly work in capacity-based application scenarios, we split the T-S application scenarios based on the network capacities between the student and the teacher.
We design three categories of T-S capacity-based application scenarios: $Cap_T > Cap_S$, $Cap_T = Cap_S$, and $Cap_T < Cap_S$, where ``T" and  ``S" denote the teacher and the student, respectively.
Specifically, we design experiments in generations in $Cap_T = Cap_S$, which means that the student in one generation is to be the teacher in the next generation.




For the competitors, KD and LT are tested in $Cap_T > Cap_S$, $Cap_T = Cap_S$, and $Cap_T < Cap_S$.
Notably, in $Cap_T = Cap_S$, the KD is employed without an ensemble, which is the same as the sequential version in \cite{Born_Again}.
Besides, in $Cap_T < Cap_S$, we implement KD by adding secondary information as the implementation in \cite{Tolerant_Teacher}.

According to \cite{Paying_more_attention,Similarity_Distillation}, AT and SP are both scenario-specialization approaches for $Cap_T > Cap_S$.
Meanwhile, the Tf-KD \cite{yuan2020revisiting} is specific to the scenario of $Cap_T = Cap_S$.

\subsection{CIFAR-10 and CIFAR-100}
We first evaluate our framework on the CIFAR-10 and CIFAR-100 datasets \cite{krizhevsky2009learning}, which consists of 60,000 32x32 RGB images in uniformly distributed 10 and 100 classes, with each class having an equal number of images.

\subsubsection{Network Baselines}

Based on the four mainstream classification models, we select nine baseline models including CNN-5, WRN-16-2, WRN-28-2, MobileNetV2, ResNet-18, WRN-16-10, ResNeXt-29, WRN-28-10, ResNet-101.
The ``ResNeXt-29" is short for ``ResNeXt-29(8 $\times$ 64d)."
The parameters of baseline networks on the CIFAR-100 are shown in Table \ref{network parameters}.



\begin{table*}[t]
  \centering
  \renewcommand\tabcolsep{17.3pt}
  \fontsize{8}{12}\selectfont
  \caption{Test error (\%) in capacity-based application scenario $Cap_T < Cap_S$ on the CIFAR-100.
  ``S(B)" and ``T(B)" are short for ``the error rate of the baseline of the student" and ``the error rate of the baseline of the teacher," respectively.}
    \begin{tabular}{|c|cc|cc|ccc|}
      \hline
      Heterogeneity & S	 & T	& S(B)   & T(B)   & KD  & LT   & \bf ICCT  \cr\hline
      \multirow{3}{*}{Architecture}
        & ResNeXt-29  & CNN-5
          & 17.70   & 39.66   & 17.58   & 17.61   & 17.48  \cr
        & ResNeXt-29  & MobileNetV2
          & 17.70   & 34.82   & 17.49   & 17.55   & 17.41  \cr
        & ResNeXt-29  & ResNet-18
          & 17.70   & 24.34   & 17.43   & 17.46 	 & 17.33 \cr\hline
      \multirow{2}{*}{Depth}
        & ResNet-101  & ResNet-18
          & 22.19  & 24.34   & 21.96   & 22.07   & 21.23  \cr
        & WRN-28-10   & WRN-16-10
          & 20.48  & 21.56   & 20.28   & 20.35   & 20.07  \cr\hline
      Width
        & WRN-28-10   & WRN-28-2
          & 20.48  & 25.28   & 20.32   & 20.38   & 20.15  \cr\hline
      Depth \& Width
        & WRN-28-10 & WRN-16-2
          & 20.48  & 27.29   & 20.37   & 20.41   & 20.26  \cr\hline
    \end{tabular}
    \label{tab:CIFAR-100_T<S}
\end{table*}

\begin{figure*}[tbp]
    \centerline{\includegraphics[width=1\textwidth]{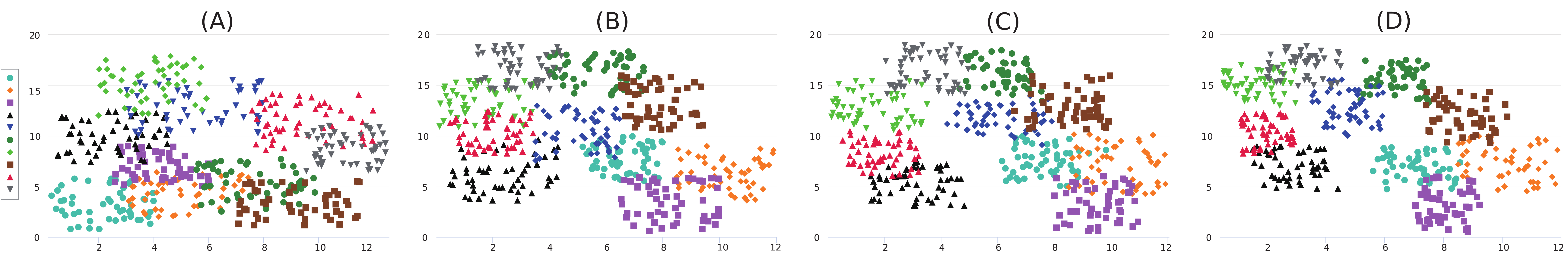}}
    \caption{
    Distributed visualization of Feature maps.
    Feature maps come from the four training methods for resnet-18 on the CIFAR-10.
    Figuer $(A)$ represents the feature map visualization of the baseline network.
    Figure $(B)$ represents the feature map visualization of the student, which is trained by our ICCT in the $Cap_T < Cap_S$ scenario.
    Figure $(C)$ represents the feature map visualization of the student, trained by our ICCT in the $Cap_T = Cap_S$ scenario.
    Figure $(D)$ represents the feature map visualization of the student, trained by our ICCT in the $Cap_T > Cap_S$ scenario.
    }
    \label{points}
\end{figure*}

\subsubsection{Training Settings}
For datasets CIFAR-10 and CIFAR-100, We applied the standard horizontal flip and random crop data augmentation.
The training protocols of CNN-5 and residual networks are not the same.
For CNN-5, we train it by Adam with batch size 256 (140 epochs; an initial learning rate of 0.001, decayed by a factor of 0.2 at epochs 40, 80, and 120).
For residual networks, We use the standard Stochastic Gradient Descent (SGD) with a weight decay of 0.0001 and a Nesterov momentum of 0.9 (200 epochs; an initial learning rate of 0.1, decayed by a factor of 0.2 at epochs 60, 120, and 160).
For KD, we set $M=4$ on the CIFAR-10, following the experiments \cite{Hinton2015DistillingTK,Similarity_Distillation,yuan2020revisiting}, and $M=10$ on the CIFAR-100.
The weights of the knowledge transfer loss are obtained by held-out validation as in \cite{Similarity_Distillation} on a subset of the training set  (CIFAR-10/100: $\lambda_{ICC} = 1500 / 1800$, $\lambda_{KD} \& \lambda_{Tf-KD} = 300 / 800$, $\lambda_{LT} = 80/150$, $\lambda_{AT} = 1000/400$ and $\lambda_{SP} = 1600 / 550$).

\subsubsection{Results and Discussions}
Following the capacity settings, we show the corresponding experimental results in the CIFAR-10 and CIFAR-100 datasets.
The part of the experimental results of CIFAR-10 is shown in the appendix.

\noindent $\bullet \ \, Cap_T > Cap_S$.


In Table \ref{tab:CIFAR-100_T>S}, the first seven T-S pairs are composed of the high-capacity teacher with large parameters and the low-capacity student with small parameters for the approach of model compression.
The last pair in this table is the high-capacity teacher with small parameters and the low-capacity student with large parameters, which is more suitable for model optimization.
Therefore, we can observe from this table, in our ICCT, no matter how many times the parameters of the teacher network differ from those of the student network, the transferring information of the teacher can always help the student improve the capability.


As shown in \ref{tab:CIFAR-100_T>S}, our ICCT can be applied in all the heterogenous T-S pairs.
However, due to the restriction of network structure in AT, AT can not handle the heterogeneous teacher and student, which are so diverse, e.g., the student of CNN-5 with the teacher of ResNet-18.
Meanwhile, the missing information of transferring information from hidden layers, e.g., the missing information of channel dims in AT, the missing information of spatial dims and spatial dims in SP, introduces a negative impact on learning results.

This table illustrates that our ICCT can systematically improve the performance of the students in the CIFAR-100 dataset.
Moreover, the error rates of students trained by ICCT are lower than other competitors regardless of the compression rate and different types of network architecture, showing the superiority of our framework.
Furthermore, due to defining the knowledge at the output layer, our ICCT is easy to extend by adding other hidden-layers based knowledge transfer techniques and obtained preferable performance.

%




\noindent $\bullet \ \, Cap_T = Cap_S$.

In this scenario, the student is taught by itself in generations.
As shown in Table \ref{tab:CIFAR-100_T=S}, the students in our ICCT perform better than themselves in all generations.
Similar to the phenomenon in \cite{Born_Again}, training models with our ICCT for multiple generations leads to inconsistent but positive improvements, which saturates after a few generations.
This improvement demonstrates that the Self-Attention based Inter-Class Correlation is useful.


\noindent $\bullet \ \, Cap_T < Cap_S$.

In this scenario, the student is taught by a low-capacity teacher.
Generally speaking, a low-capacity teacher can not teach the high-capacity student well.
However, the results in Table \ref{tab:CIFAR-100_T<S} show that the high-capacity model can also learn something useful from the low-capacity model.
Though the low-capacity teacher transfers a gentler supervision signal, the Inter-Class Correlation from the teacher can serve as a generic regularization
method and reduce the risk of overfitting for the high-capacity student.

To further analyze the effectiveness of our ICCT in three kinds of capacity-based scenarios, we conduct two-dimensional visualization of the feature map distribution after the fully connected layer of Resnet-18.
Following CIFAR-10 experiments, we can clearly observe from Figure \ref{points}, compared with all students in our ICCT, the feature map distribution of the baseline is significantly disordered.
In contrast, our ICCT devotes to cluster the feature map by reducing the inner-class distance and expanding the inter-class distance.
In the three scenarios, the student taught by the teachers with high-capacity, learns more compact and discriminative representations.
Besides, there are encouraging signs that even the low-capacity teacher still has certain help for the student to obtain the better capability of feature understanding.

\begin{table*}[t]
  \centering
  \renewcommand\tabcolsep{7.3pt}
  \fontsize{8}{12}\selectfont
  \caption{Test error (\%) in capacity-based application scenario $Cap_T > Cap_S$ on the ILSVRC2012.
  ``S(B)" and ``T(B)" are short for ``the error rate of the baseline of the student" and ``the error rate of the baseline of the teacher," respectively.}
    \begin{tabular}{|c|cc|cc|cccc|ccc|}
    \hline
      Heterogeneity & S  & T  & S(B) & T(B)  & KD   & LT   & AT   & SP & \bf ICCT   & \bf ICCT+AT & \bf ICCT+SP  \cr\hline
      \multirow{3}{*}{Architecture}
            & ResNet-18 & ResNeXt-101    & 29.12  & 20.40
              & 28.74  & 28.94  & 28. 31  & 28.02  & 27.66  & 27.52 & 27.41 \cr
            & MobileNetV2 & ResNeXt-101  & 28.13   & 20.40
              & 27.90  & 27.97  & 27.43  & 27.12   & 26.74  & 26.63	& 26.45  \cr
            & ResNet-50 & ResNeXt-101    & 22.71   & 20.40
              & 22.16  & 22.32  & 21.87  & 21.75   & 21.22  & 21.03 & 20.86 \cr\hline
      \multirow{2}{*}{Depth}
            & ResNet-18 & ResNet-50  & 29.12    & 22.71
              & 28.78  & 28.94  & 28.65  & 28.42  & 28.19  & 27.86  & 27.73 \cr
            & WRN-18-2  & WRN-34-2   & 25.56    & 23.35
              & 25.28  & 25.30  & 25.14  & 25.03 & 24.81  & 24.62  & 24.47  \cr\hline
      Width
            & WRN-34-1  & WRN-34-2   & 26.74    & 23.35
              & 26.37  & 26.45  & 26.32  & 26.18  & 25.86  & 25.79  & 25.56  \cr\hline
      Depth \& Width
            & WRN-18-1  & WRN-34-2   & 30.38    & 23.35
              & 29.96  & 30.08  & 29.71  & 29.57  & 29.32  & 29.25  & 29.04 \cr\hline
    \end{tabular}
    \label{tab:ImageNet_T>S}
\end{table*}

\begin{table}[t]
  \centering
  \renewcommand\tabcolsep{6.3pt}
  \fontsize{8}{12}\selectfont
  \caption{Test error (\%) in capacity-based application scenario $Cap_T = Cap_S$ on the ILSVRC2012.}
    \begin{tabular}{|c|c|cccc|}
    \hline
      Name      & Generation  &  KD  & LT  & Tf-KD  & \bf ICCT    \cr\hline
      \multirow{5}{*}{ResNet-18}
                &  Baseline	 & 29.12 	&	29.12 	&	29.12	 & 29.12  \\ \cline{2-6}
                &  Gen \#1	 & 29.03  &	29.08 	&	28.84  & 28.91  \cr
                &  Gen \#2	 & 28.94	& 29.13 	&	28.75  & 28.66  \cr
                &  Gen \#3	 & 28.91  &	28.96 	&	28.83  & 28.25  \cr
                &  Gen \#4	 & 28.97  &	29.04 	&	28.72  & 28.41  \cr\hline
      \multirow{5}{*}{MobileNetV2}
                &  Baseline	 & 28.13  &	28.13 	&	28.13	 & 28.13  \\ \cline{2-6}
                &  Gen \#1	 & 27.97  &	28.15 	&	27.89  & 26.86  \cr
                &  Gen \#2	 & 28.03	& 28.02 	&	27.54  & 26.77  \cr
                &  Gen \#3	 & 27.94  &	28.06 	&	26.97  & 27.50  \cr
                &  Gen \#4	 & 27.96  & 28.03  	&	27.15  & 26.91  \cr\hline
    \end{tabular}
    \label{tab:ImageNet_T=S}
\end{table}

\begin{table*}[t]
  \centering
  \renewcommand\tabcolsep{15.8pt}
  \fontsize{8}{12}\selectfont
  \caption{Test error (\%) in capacity-based application scenario $Cap_T < Cap_S$ on the ILSVRC2012.
  ``S(B)" and ``T(B)" are short for ``the error rate of the baseline of the student" and ``the error rate of the baseline of the teacher," respectively.}
    \begin{tabular}{|c|cc|cc|ccc|}
    \hline
      Heterogeneity & S  & T  & S(B) & T(B)   & KD  & LT  & \bf ICCT     \cr\hline
      \multirow{3}{*}{Architecture}
            & ResNeXt-101  & ResNet-18   & 20.40    & 29.12
              & 20.26      & 20.34       & 20.15    \cr
            & ResNeXt-101  & MobileNetV2 & 20.40    & 28.13
              & 20.21      & 20.28       & 20.03    \cr
            & ResNeXt-101  & resnet-50   & 20.40    & 22.71
              & 20.16      & 20.25       & 19.97    \cr\hline
      \multirow{2}{*}{Depth}
            & resnet-50    & ResNet-18   & 22.71    & 29.12
              & 22.44      & 22.56       & 22.10    \cr
            & WRN-34-2     & WRN-18-2    & 23.35    & 25.56
              & 23.08      & 23.14       & 22.95    \cr\hline
      Width
            & WRN-34-2     & WRN-34-1    & 23.35    & 26.74
              & 23.19      & 23.21       & 23.13    \cr\hline
      Depth \& Width
            & WRN-34-2     & WRN-18-1    & 23.35    & 30.38
              & 23.22      & 23.24       & 23.18    \cr\hline
    \end{tabular}
    \label{tab:ImageNet_T<S}
\end{table*}

\subsection{ILSVRC2012}
We now investigate the more complex ILSVRC2012 dataset \cite{russakovsky2015imagenet}.
The ILSVRC2012 dataset is a popular subset of the ImageNet database \cite{deng2009imagenet}, which contains 1.3 million high-resolution images and 1,000 classes in total.

\subsubsection{Network Baselines}
To handle the ILSVRC2012 dataset, which is more complicated than CIFAR-10 and CIFAR-100, we choose the models with higher capacity and leave alone the CNN-5 model.

The baselines contain ResNet-18, ResNet-50, ResNeXt-101 (64 $\times$ 4d), MobileNetV2, WRN-18-1, WRN-18-2, WRN-34-1, WRN-34-2
We abbreviate ResNeXt-101 (64 $\times$ 4d) as ResNeXt-101.


%
%
%
%
%
%

\subsubsection{Training Settings}
Due to the complex of the ILSVRC2012 dataset, a series of data-augmentation techniques are applied, including cropping, rescaling, and randomly mirroring the image.
For all networks, The SGD with a weight decay of 0.0001 and a Nesterov momentum of 0.9 is used with batch size 256 (90 epochs; an initial learning rate of 0.1, decayed by the cosine annealing).
For KD, we fix $M=5$, following the experiments in \cite{Hinton2015DistillingTK,Tolerant_Teacher,yuan2020revisiting}.
The weights of the knowledge transfer loss are obtained as in \cite{Similarity_Distillation} on a subset of the training set ($\lambda_{ICC} = 2000$, $\lambda_{KD} \& \lambda_{Tf-KD} = 350$, $\lambda_{LT} = 100$,  $\lambda_{AT} = 120$ and $\lambda_{SP} = 850$).

\subsubsection{Results and Discussions}
The results are also presented in three capacity-based application scenarios.

\noindent $\bullet \ \, Cap_T > Cap_S$.

   Due to the complexity of the task, though the advantage of the ICCT in the ILSVRC2012 dataset is not as much as that in the CIFAR-10 and CIFAR-100 datasets, the ICCT still outperforms other competitors.
The possible reason is that the teacher is also unable to completely handle the complex task, resulting in the teacher's ICC containing less useful information.

\noindent $\bullet \ \, Cap_T = Cap_S$.

As shown in Table \ref{tab:ImageNet_T=S}, the improvement from our ICCT is also inconsistent from generation to generation, which is similar to previous CIFAR-10 and CIFAR-100 experiments.
The third generation of ResNet-18 in our ICCT presents the test rate of $28.25\%$, which is very close to the same student taught by ResNet-50 $28.19\%$.
This small gap demonstrates the potential of training neural networks in generations.

\noindent $\bullet \ \, Cap_T < Cap_S$.

We can observe similar results in Table \ref{tab:ImageNet_T<S} as those in the CIFAR-100 experiments.
The ICCT can bring in relatively little but consistent progress in different students and teachers.
Simultaneously, our ICCT still outperforms other competitors.
This experiment proves that a low-capacity teacher can still improve the performance of high-capacity students through the ICCT in complicated tasks.

\section{Related Work}
\subsection{Teacher-Student Framework}
The goal of the T-S framework is training a student model by learning from the knowledge of the teacher model to achieve better generalizing accuracy than being trained directly.
The performance of the T-S framework is sensitive to how knowledge is defined.

Some early researches defined the knowledge at the output layer by extracting something useful in the outputs.
Hinton \emph{et al.} \cite{Hinton2015DistillingTK} first proposed the Knowledge Distillation (KD) by introducing the concept of ``dark knowledge," which is the softened output of the teacher's softmax distribution with a soften hyperparameter ``temperature."
KD can be regarded as a generalization of previous work \cite{ba2014deep} by setting the ``temperature" coefficient large enough.

After that, some studies turned to define the knowledge in the hidden layers, aiming at teaching the student step by step.
Zagoruyko and Komodakis \cite{Paying_more_attention} forced the student to match the attention map of the teacher (norm across the channel dimension in each spatial location) at the end of each residual stage.
Czarnecki \emph{et al.} \cite{czarnecki2017sobolev} tried to minimize the difference between the teacher and student derivatives of the loss concerning the input.
Yim \emph{et al.} \cite{A_Gift} designed a correlation matrix of two hidden layers' feature maps to be the knowledge and employed Knowledge Distillation (KD) in the final layer.
Tung and Mori \cite{Similarity_Distillation} and Hou \emph{et al.} \cite{Lane_Detection} defined the knowledge by the similarity activation maps in hidden layers.
Excepting \cite{Similarity_Distillation}, the output shape of student's hidden layers should be similar to the teachers', considering the complexity of hidden features.

In the image classification task, all the above frameworks were implemented in the application scenario of $Cap_{T}>Cap_{S}$.
Recently, the potential of the T-S framework has been further explored.
Based on the model capacity of the teacher ($Cap_{T}$) and the student ($Cap_{S}$), the capacity-based application scenarios of T-S frameworks could be divided into three categories: (1) $Cap_{T}>Cap_{S}$; (2) $Cap_{T}=Cap_{S}$; (3) $Cap_{T}<Cap_{S}$.
Furlanello \emph{et al.} \cite{Born_Again} and Yuan \emph{et al.} \cite{yuan2020revisiting} considered the scenario $Cap_{T}=Cap_{S}$ and demonstrated the capability of KD to improve the performance of a network by itself in generations.
Yang \emph{et al.} \cite{Tolerant_Teacher} and Yuan \emph{et al.} \cite{yuan2020revisiting} researched the more difficult scenario $Cap_{T}<Cap_{S}$ and find that the student's accuracy could also be improved by secondary information and KD in generations, although the teacher has a lower capacity than the student.


Another important question is how to transfer the knowledge between students and teachers.
Inspired by generative adversarial networks, Wang \emph{et al.} \cite{KDGAN} introduced an additional discriminator network to the original T-S framework, and jointly trained the student, teacher, and discriminator networks.
Other ensemble-based variants, such as Lan \emph{et al.} \cite{Knowledge_ensemble}, utilized the techniques of ensemble learning to train a teacher together with multiple students.

\section{Conclusion and Future Work}
In this paper, we study the problem of improving the performance of a student neural network by a teacher neural network in the classification task.
To tackle this problem, we propose a new Teacher-Student framework named Inter-Class Correlation Transfer (ICCT), in which the transferring knowledge is defined as the ICC map implemented by the Self-Attention mechanism.
ICC map eliminates the restrictions of network structure between the teacher and the student, and avoids the loss of beneficial transferring information.
Our ICCT works in a ``comprehensive" mode, which means the student can jointly utilize the fine-grained pairwise correlations from the teacher and add its own beliefs.
Since the ``comprehensive" mode provides an effective regularization, our ICCT can be flexibly implemented in all capacity-based T-S application scenarios.
Experiments validate the availability of our ICCT and show that it is superior to existing frameworks in all capacity-based scenarios, $Cap_T > Cap_S$, $Cap_T = Cap_S$, and $Cap_T < Cap_S$, with different network structures, regardless of the complexity of the task.
Moreover, due to defining the knowledge at the output layer, our ICCT is easy to extend by adding other hidden-layers based knowledge transfer techniques.

In the future, we will study other forms of ICC besides Self-Attention and extend the pairwise correlations to tripartite or more correlations.
We believe the performance of ICCT can be further promoted with additional Inter-Class Correlation.
Furthermore, we will consider the inter-sample correlations to verify whether they can be useful to transfer.
This may be helpful to the research of the random crop strategies.

{\small
\bibliographystyle{IEEEtran}
\bibliography{egbib}

\begin{thebibliography}{10}
\providecommand{\url}[1]{#1}
\csname url@samestyle\endcsname
\providecommand{\newblock}{\relax}
\providecommand{\bibinfo}[2]{#2}
\providecommand{\BIBentrySTDinterwordspacing}{\spaceskip=0pt\relax}
\providecommand{\BIBentryALTinterwordstretchfactor}{4}
\providecommand{\BIBentryALTinterwordspacing}{\spaceskip=\fontdimen2\font plus
\BIBentryALTinterwordstretchfactor\fontdimen3\font minus
  \fontdimen4\font\relax}
\providecommand{\BIBforeignlanguage}[2]{{%
\expandafter\ifx\csname l@#1\endcsname\relax
\typeout{** WARNING: IEEEtran.bst: No hyphenation pattern has been}%
\typeout{** loaded for the language `#1'. Using the pattern for}%
\typeout{** the default language instead.}%
\else
\language=\csname l@#1\endcsname
\fi
#2}}
\providecommand{\BIBdecl}{\relax}
\BIBdecl

\bibitem{Large_Networks}
N.~Shazeer, A.~Mirhoseini, K.~Maziarz, A.~Davis, Q.~Le, G.~Hinton, and J.~Dean,
  ``Outrageously large neural networks: The sparsely-gated mixture-of-experts
  layer,'' in \emph{Proceedings of the International Conference on Learning
  Representations}, 2017.

\bibitem{krizhevsky2012imagenet}
A.~Krizhevsky, I.~Sutskever, and G.~H. Hinton, ``Imagenet classification with
  deep convolutional neural networks,'' in \emph{Proceedings of the Advances in
  Neural Information Processing Systems}, 2012.

\bibitem{Inception-V4}
C.~Szegedy, S.~Ioffe, V.~Vanhoucke, and A.~A. Alemi, ``Inception-v4,
  inception-resnet and the impact of residual connections on learning,'' in
  \emph{Proceedings of the Thirty-First AAAI Conference on Artificial
  Intelligence}, 2017.

\bibitem{Multigrid}
T.-W. Ke, M.~Maire, and S.~X. Yu, ``Multigrid neural architectures,'' in
  \emph{Proceedings of the IEEE Conference on Computer Vision and Pattern
  Recognition}, 2017.

\bibitem{bucilua2006model}
C.~Buciluǎ, R.~Caruana, and A.~Niculescu-Mizil, ``Model compression,'' in
  \emph{Proceedings of the ACM SIGKDD on Knowledge Discovery and Data Mining},
  2006.

\bibitem{he2016deep}
K.~He, X.~Zhang, S.~Ren, and J.~Sun, ``Deep residual learning for image
  recognition,'' in \emph{Proceedings of the IEEE Conference on Computer Vision
  and Pattern Recognition}, 2016.

\bibitem{model_compression}
Y.~Cheng, D.~Wang, P.~Zhou, and T.~Zhang, ``A survey of model compression and
  acceleration for deep neural networks,'' \emph{IEEE SIgnal Processing
  Magazine, Special issue on Deep Learning for IMAGE understanding}, 2017.

\bibitem{Hinton2015DistillingTK}
G.~Hinton, O.~Vinyals, and J.~Dean, ``Distilling the knowledge in a neural
  network,'' \emph{Proceedings of the Advances in Neural Information Processing
  Systems}, 2015.

\bibitem{heo2019comprehensive}
B.~Heo, J.~Kim, S.~Yun, H.~Park, N.~Kwak, and J.~Y. Choi, ``A comprehensive
  overhaul of feature distillation,'' in \emph{Proceedings of the IEEE
  International Conference on Computer Vision}, 2019.

\bibitem{Similarity_Distillation}
F.~Tung and G.~Mori, ``Similarity-preserving knowledge distillation,'' in
  \emph{Proceedings of the IEEE International Conference on Computer Vision},
  2019.

\bibitem{Born_Again}
T.~Furlanello, Z.~C. Lipton, M.~Tschannen, L.~Itti, and A.~Anandkumar, ``Born
  again neural networks,'' in \emph{Proceedings of the International Conference
  on Machine Learning}, 2018.

\bibitem{yuan2020revisiting}
L.~Yuan, F.~E. Tay, G.~Li, T.~Wang, and J.~Feng, ``Revisiting knowledge
  distillation via label smoothing regularization,'' in \emph{Proceedings of
  the IEEE Conference on Computer Vision and Pattern Recognition}, 2020.

\bibitem{Tolerant_Teacher}
C.~Yang, L.~Xie, S.~Qiao, and A.~Yuille, ``Training deep neural networks in
  generations: A more tolerant teacher educates better students,'' in
  \emph{Proceedings of the AAAI Conference on Artificial Intelligence}, 2019.

\bibitem{Paying_more_attention}
S.~Zagoruyko and N.~Komodakis, ``{Paying more attention to attention: improving
  the performance of convolutional neural networks via attention transfer},''
  in \emph{Proceedings of the International Conference on Learning
  Representations}, 2017.

\bibitem{A_Gift}
J.~Yim, D.~Joo, J.~Bae, and J.~Kim, ``A gift from knowledge distillation: Fast
  optimization, network minimization and transfer learning,'' in
  \emph{Proceedings of the IEEE Conference on Computer Vision and Pattern
  Recognition}, 2017.

\bibitem{Lane_Detection}
Y.~Hou, Z.~Ma, C.~Liu, and C.~C. Loy, ``Learning lightweight lane detection
  cnns by self attention distillation,'' in \emph{IEEE International Conference
  on Computer Vision}, 2019.

\bibitem{vaswani2017attention}
A.~Vaswani, N.~Shazeer, N.~Parmar, J.~Uszkoreit, L.~Jones, A.~N. Gomez,
  {\L}.~Kaiser, and I.~Polosukhin, ``Attention is all you need,'' in
  \emph{Proceedings of the Advances in Neural Information Processing Systems},
  2017.

\bibitem{krizhevsky2009learning}
A.~Krizhevsky and G.~Hinton, ``Learning multiple layers of features from tiny
  images,'' 2009.

\bibitem{russakovsky2015imagenet}
O.~Russakovsky, J.~Deng, H.~Su, J.~Krause, S.~Satheesh, S.~Ma, Z.~Huang,
  A.~Karpathy, A.~Khosla, M.~Bernstein, A.~C. Berg, and L.~Fei-Fei, ``Imagenet
  large scale visual recognition challenge,'' \emph{International Journal of
  Computer Vision}, 2015.

\bibitem{wen2016discriminative}
Y.~Wen, K.~Zhang, Z.~Li, and Y.~Qiao, ``A discriminative feature learning
  approach for deep face recognition,'' in \emph{Proceedings of the European
  Conference on Computer Vision}, 2016.

\bibitem{bengio2013using}
S.~Bengio, J.~Dean, D.~Erhan, E.~Ie, Q.~Le, A.~Rabinovich, J.~Shlens, and
  Y.~Singer, ``Using web co-occurrence statistics for improving image
  categorization,'' \emph{arXiv preprint arXiv:1312.5697}, 2013.

\bibitem{rabinovich2007objects}
A.~Rabinovich, A.~Vedaldi, C.~Galleguillos, E.~Wiewiora, and S.~B. Belongie,
  ``Objects in context,'' in \emph{Proceedings of the IEEE International
  Conference on Computer Vision}, 2007.

\bibitem{wu2014exploring}
Z.~Wu, Y.-G. Jiang, J.~Wang, J.~Pu, and X.~Xue, ``Exploring inter-feature and
  inter-class relationships with deep neural networks for video
  classification,'' in \emph{Proceedings of the ACM International Conference on
  Multimedia}, 2014.

\bibitem{cho2014learning}
K.~Cho, B.~van Merrienboer, C.~Gulcehre, D.~Bahdanau, F.~Bougares, H.~Schwenk,
  and Y.~Bengio, ``Learning phrase representations using rnn encoder--decoder
  for statistical machine translation,'' in \emph{Proceedings of the Empirical
  Methods in Natural Language Processing}, 2014.

\bibitem{paulus2017deep}
A.~P. Parikh, O.~T{\"a}ckstr{\"o}m, D.~Das, and J.~Uszkoreit, ``A decomposable
  attention model for natural language inference,'' in \emph{Proceedings of the
  International Conference on Learning Representations}, 2018.

\bibitem{bahdanau2014neural}
D.~Bahdanau, K.~Cho, and Y.~Bengio, ``Neural machine translation by jointly
  learning to align and translate,'' in \emph{Proceedings of the International
  Conference on Learning Representations}, 2015.

\bibitem{ba2014deep}
L.~J. Ba and R.~Caurana, ``Do deep nets really need to be deep?'' in
  \emph{Proceedings of the Advances in Neural Information Processing Systems},
  2014.

\bibitem{xie2017aggregated}
S.~Xie, R.~Girshick, P.~Doll{\'a}r, Z.~Tu, and K.~He, ``Aggregated residual
  transformations for deep neural networks,'' in \emph{Proceedings of the IEEE
  Conference on Computer Vision and Pattern Recognition}, 2017.

\bibitem{WideResNet}
S.~Zagoruyko and N.~Komodakis, ``Wide residual networks,'' in \emph{Proceedings
  of the British Machine Vision Conference, 2016}, 2016.

\bibitem{mobilenetv2}
M.~Sandler, A.~Howard, M.~Zhu, A.~Zhmoginov, and L.-C. Chen, ``Mobilenetv2:
  Inverted residuals and linear bottlenecks,'' in \emph{Proceedings of the IEEE
  Conference on Computer Vision and Pattern Recognition}, 2018.

\bibitem{deng2009imagenet}
J.~Deng, W.~Dong, R.~Socher, L.-J. Li, K.~Li, and F.~F. Li, ``Imagenet: A
  large-scale hierarchical image database,'' in \emph{Proceedings of the IEEE
  Conference on Computer Vision and Pattern Recognition}, 2009.

\bibitem{czarnecki2017sobolev}
W.~M. Czarnecki, S.~Osindero, M.~Jaderberg, G.~Swirszcz, and R.~Pascanu,
  ``Sobolev training for neural networks,'' in \emph{Proceedings of the
  Advances in Neural Information Processing Systems}, 2017.

\bibitem{KDGAN}
X.~Wang, R.~Zhang, Y.~Sun, and J.~Qi, ``Kdgan: knowledge distillation with
  generative adversarial networks,'' in \emph{Proceedings of the Advances in
  Neural Information Processing Systems}, 2018.

\bibitem{Knowledge_ensemble}
X.~Lan, X.~Zhu, and S.~Gong, ``Knowledge distillation by on-the-fly native
  ensemble,'' in \emph{Proceedings of the Advances in Neural Information
  Processing Systems}, 2018.

\end{thebibliography}
}

\end{document}